\pdfoutput=1

\documentclass[11pt]{article}

\usepackage[preprint]{acl}

\usepackage{times}
\usepackage{latexsym}

\usepackage{booktabs}
\usepackage{url}
\usepackage{hyperref}
\usepackage{adjustbox}
\usepackage{multirow}
\usepackage{float}
\usepackage{subcaption}

\usepackage[T1]{fontenc}

\usepackage[utf8]{inputenc}

\usepackage{microtype}

\usepackage{inconsolata}

\usepackage{graphicx}

\usepackage{amsmath} 
\DeclareFontEncoding{LS1}{}{}
\DeclareFontSubstitution{LS1}{stix}{m}{n}
\DeclareRobustCommand{\mathvisiblespace}{%
  \mathord{\text{\usefont{LS1}{stixscr}{m}{n}\symbol{"B6}}}%
}

\title{Subword models struggle with word learning, but surprisal hides it}

\author{Bastian Bunzeck \and Sina Zarrieß \\
  Computational Linguistics, Department of Linguistics \\
  Bielefeld University, Germany \\
  \texttt{\{bastian.bunzeck, sina.zarriess\}@uni-bielefeld.de}}

\begin{document}
\maketitle
\begin{abstract}
We study word learning in subword and character language models with the psycholinguistic lexical decision task. While subword LMs struggle to discern words and non-words with high accuracy, character LMs solve this task easily and consistently. Only when supplied with further contexts do subword LMs perform similarly to character models. Additionally, when looking at word-level and syntactic learning trajectories, we find that both processes are separable in character LMs. Word learning happens before syntactic learning, whereas both occur simultaneously in subword LMs. This raises questions about the adequacy of subword LMs for modeling language acquisition and positions character LMs as a viable alternative to study processes below the syntactic level.
\end{abstract}

\section{Introduction}

When humans acquire their first language(s), they \textit{first} learn to recognize single words, mostly from short, fragmentary utterances \cite{cameron-faulkner2003construction,bunzeck2024richness}, 
\textit{before} fully understanding the grammatical processes governing them \cite{tomasello1992first,behrens2021constructivist}. This simple fact about language acquisition has received surprisingly little attention in the body of work that treats LMs as models of language learners \cite{warstadt2022what,portelance2024roles}. While word learning in children is comparatively well studied \citep{plunkett1997theories,yu2007unified,waxman2009early,bergelson201269,clark2015first,frank2021variability}, the implicit word learning processes in LMs are not. Current studies focus on syntax \cite{mueller2022coloring,choshen2022grammarlearning}, or investigate word learning in close connection with syntax through surprisal \cite{chang2022word,portelance2023predicting,shafiabadi2025surprisal,ficarra2025distributional}. Architecture-wise, a key limitation to the precise study of word learning is subword tokenization (e.g., BPE, \citealp{gage1994new}), which splits words into linguistically \cite{arnett2025why} and cognitively implausible units \cite{beinborn2023analyzing}.

\begin{figure}[t!]
\centering
\includegraphics[width=\linewidth]{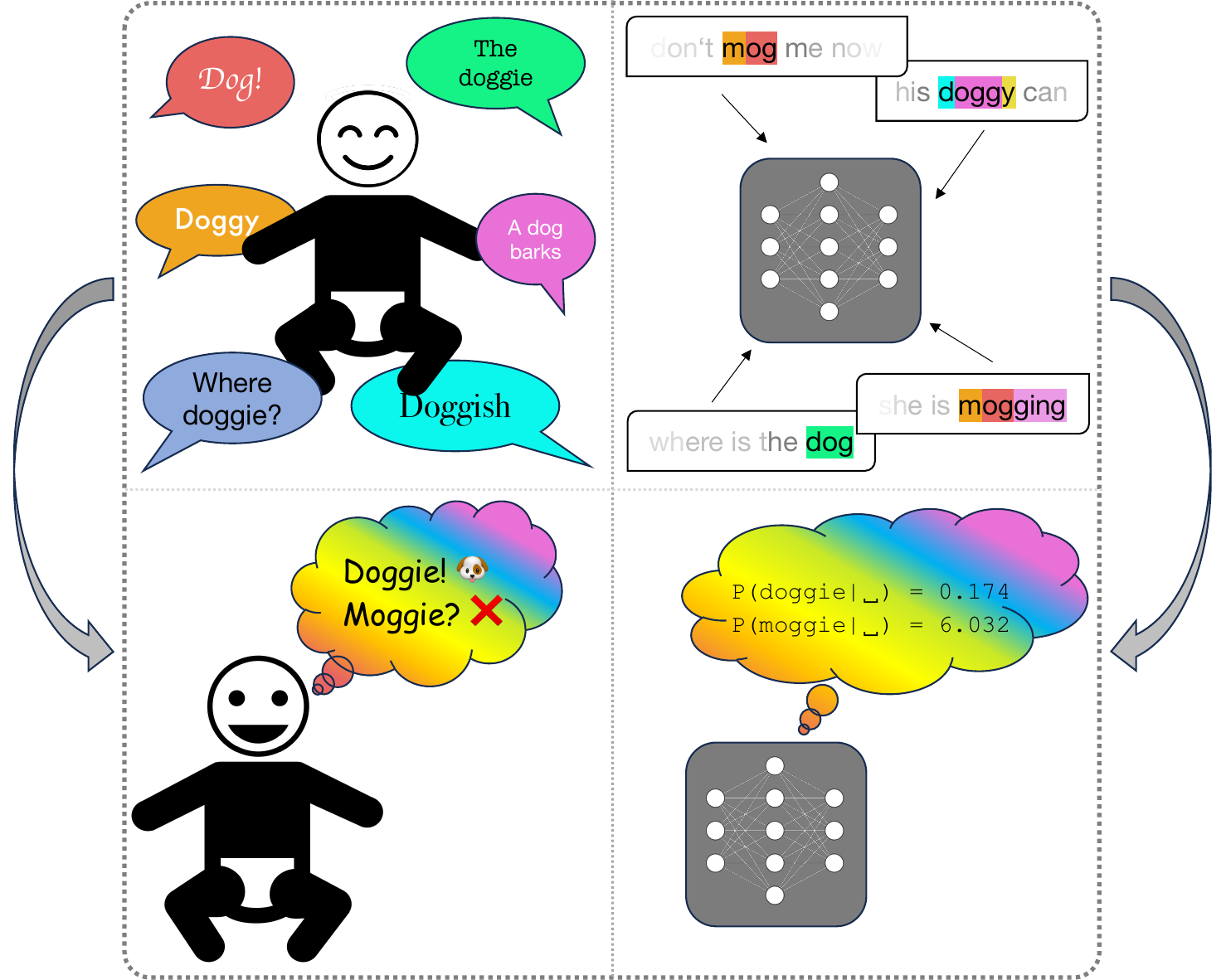}
\vspace{-0.5cm} 
\caption{Illustration of word learning in human learners and transformer LLMs (top), and of our lexical decision test that probes discrimination of words from non-words (bottom). While human learners build up an mental lexicon from experience with language, artificial learners assign probabilities to strings based on their frequency.}
\vspace{-0.6cm} 
\label{fig:bla}
\end{figure}

To gauge word learning in a syntax-independent manner, we use the psycholinguistic lexical decision task \cite{meyer1971facilitation,legodais2017comparing}, i.e., deciding which word in a given word/non-word pair is real. We find that models with character-level tokenization learn this task quickly and reliably. In contrast, subword LMs of all sizes perform significantly worse in a syntax-independent setting and only achieve comparable accuracy when stimuli are measured through surprisal, or ``unexpectedness,'' in linguistic context. By comparing word and syntactic learning (measured via BLiMP, \citealp{warstadt2020blimp}), we further find that character models quickly acquire word knowledge and only later develop syntactic knowledge. In subword models, word and syntax learning happen concurrently. This demonstrates how elementary modeling decisions, such as tokenization methods, significantly impact learning trajectories in LMs, a fact that warrants more scrutiny when using LMs as models of language acquisition.

\section{Related work} \label{sec:relwork}

Word learning in humans is a multifaceted phenomenon that involves different kinds of (extra)linguistic knowledge \cite{waxman2009early}: while phoneticians are concerned with word recognition and sequence segmentation \cite{jusczyk1999how}, developmental psychologists frequently equate word learning with correct reference to real world objects \cite{stager1997infants,ackermann2020children}. Psycholinguists usually focus on the mental lexicon and learning which words belong to it \cite{goldinger1996auditory}, whereas usage-based scholars take into account children's productions and their ability to be competent language users, even with few words \cite{tomasello1992first}.

Although aspects like word recognition and sequence segmentation have been studied in LMs (e.g. \citealp{goriely2025babylms}), the most common approach to word learning in LMs is measuring the predictability of words via surprisal (negative log-probability, \citealp{hale2001probabilistic}). \citet{chang2022word} train LMs on book texts and wiki data. They define a surprisal threshold below which words are said to be learned and find that frequent function words are learned earliest. Here, the alignment between models and real learners is questionable: children first utter nouns and verbs \cite{tomasello2000itembased}, but also rely on function words for challenges like speech segmentation \cite{dye2019lexical}. \citet{portelance2023predicting} show that in LSTMs trained on child-directed speech, surprisal correlates with word-level age of acquisition. \citet{chang2024characterizing} observe that learning curves for surprisal values are stable for frequent tokens, while infrequent tokens are ``forgotten'' again over pre-training. \citet{shafiabadi2025surprisal} introduce anti-surprisal (in incorrect contexts) to track false usage, which also fluctuates over pre-training. These studies cast word learning as the ability to anticipate words' expectedness in a given syntactic (and semantic) context. We note a certain conceptual leap to the original works on surprisal, where it is primarily viewed as an incremental measure of processing difficulty in syntactic comprehension \cite{levy2008expectationbased, demberg2009computational}. A simple word like \textit{dog} might be surprising and therefore hard to parse in some contexts, but very expected in others, independently of being already learned on the pure word level. A further methodological drawback of surprisal as a measure of word learning is that it corresponds almost directly to the next-token prediction objective LMs are trained on. This contrasts with typical probing paradigms used in the domain of syntax, which implement the idea to ``challenge'' models in minimal pair set-ups that are not observed directly as string sequences in training, thereby testing abstracted, implicit linguistic knowledge rather than observed patterns in the data. In a similar vein, we want to probe the word knowledge of an LM at a fundamental level and beyond surface-level word sequences that LMs are known to excel in predicting. We want to know if the artificial learner knows that the word \textit{doggie} exists in the English language, but \textit{moggie} does not. 

Lexical decision is widely used in human studies but remains an underexplored LM benchmark. \citet{legodais2017comparing} show that character-based LSTMs achieve about 95\% accuracy on such tasks. \citet{lavechin2023babyslm} find that speech-based LMs need significantly more input to still perform poorly (56.8\%) than phoneme-level LSTMs (75.4\%) on a phonetic dataset. For the same data,  \citet{goriely2024babble} find that GPT-2-based subword BabyLMs achieve 70\% accuracy and a comparable character models reach nearly 90\%. Finally, for another lexical decision dataset, \citet{bunzeck2025small} report near-perfect  accuracy for character-based grapheme Llama models, while phoneme models perform at 60--70\%. 

\begin{table*}[t!]
\footnotesize
\centering
\begin{tabular}{@{}clrc|cc|cc|cc@{}}
\toprule
 &  & &  & \multicolumn{2}{c}{Lexical decision} & \multicolumn{2}{c}{Surprisal} & \multicolumn{2}{c}{Anti-surprisal} \\ 
Tokenization & Model & Parameters & Data size & highFrq & lowFrq & highFrq & lowFrq & highFrq & lowFrq \\ \midrule
\multirow{10}{*}{\rotatebox[origin=c]{90}{Subword (BPE)}} & \multirow{6}{*}{Pythia} & 14M & \multirow{6}{*}{825GB} & 66.6 & 62.5 & 90.5 & 85.5 & 71.4 & 77.7 \\
 &  & 70M & & 72.5 & 68.8 & 94.5 & 94.0 & 77.0 & 83.6 \\
 &  & 160M & & 77.8 & 73.0 & 96.4 & 95.8 & 78.0 & 85.7 \\
 &  & 410M & & 81.9 & 78.1 & 97.7 & 97.9 & 77.1 & 84.1 \\
 &  & 1B & & 87.5 & 83.2 & 97.7 & 97.9 & 76.6 & 83.8 \\
 &  & 1.4B & & 87.8 & 81.6 & 97.9 & 97.9 & 76.5 & 84.7 \\ \cmidrule(l){2-10}
 & GPT-2 & 97.5M & 100M words & 35.6 & 79.1 & 99.0 & 99.2 & 84.7 & 86.9 \\ \cmidrule(l){2-10}
 & \multirow{3}{*}{Llama} & 2.51M & \multirow{3}{*}{10M words} & 70.9 & 58.4 & 86.7 & 70.9 & 78.6 & 67.7 \\
 &  & 7.77M & & 79.5 & 63.2 & 91.3 & 78.1 & 81.1 & 72.9 \\
 &  & 30.03M & & 83.6 & 68.6 & 92.7 & 81.1 & 83.7 & 76.1 \\ \midrule
\multirow{4}{*}{\rotatebox[origin=c]{90}{Character}} & GPT-2 & 85.3M & 100M words & 98.7 & \textbf{97.3} & \textbf{99.8} & \textbf{99.4} & 98.0 & \textbf{96.3} \\ \cmidrule(l){2-10}
 & \multirow{3}{*}{Llama} & 0.49M & \multirow{3}{*}{10M words} & 97.6 & 83.0 & 98.2 & 84.3 & 98.0 & 83.1 \\
 &  & 3.73M & & 98.9 & 90.2 & 99.4 & 90.3 & 98.5 & 88.8 \\
 &  & 21.94M & & \textbf{99.0} & 93.3 & \textbf{99.8} & 94.7 & \textbf{99.0} & 92.5 \\ \midrule
\end{tabular}
\vspace{-0.2cm} 
\caption{Accuracy scores (in \%) for (i) lexical decision, (ii) surprisal and (iii) anti-surprisal experiments}
\vspace{-0.5cm} 
\label{tab:results-exp12}
\end{table*}

\section{Experiments}

\paragraph{Models}
We train triplets of increasingly larger Llama models \cite{touvron2023llama} with character/subword tokenization on the BabyLM 10M corpus \cite{choshen2024call}. Training details are found in Appendix \ref{sec:model-stuff}. As ablations, we test subword Pythias \cite{biderman2023pythia} and character/subword GPT-2 models \citep{goriely2024babble}.

\paragraph{Test data}
We follow the idea of forced-choice lexical decision \cite{baddeley1993spottheword}, where participants must decide which is real: an existing word or a synthesized non-word. We use \texttt{wuggy} \cite{keuleers2010wuggy} to generate minimal pairs of words/non-words that differ in one or two syllables, akin to syntactic minimal pair tests such as BLiMP. We derive 1,000 non-words (e.g. \textit{monding}) each from 1,000 high-frequency/low-frequency words (e.g. \textit{sending}), which preserve syllable-bigram frequencies and match their origin words in length (cf. Appendix \ref{sec:tokenization}). 

\paragraph{Lexical decision} 
For a word/non-word pair $(w,*w)$, we measure $-log(P(w|\mathvisiblespace))$ and $-log(P(*w|\mathvisiblespace))$, i.e. how ``surprised'' a LM is by the word in the context of a prepended whitespace (and BOS token). If $-log(P(w|\mathvisiblespace)) < -log(P(*w|\mathvisiblespace))$, the LM's lexical decision is correct. As autoregressive LMs are sequence prediction models, we need a preceding context for which we can calculate surprisal. A single whitespace is the most neutral starting token available (and for subword models also signals that the first subword is word-initial). For all experiments, we calculate the average surprisal over all tokens of a word (which, in some cases, is characterized by a mismatch in token numbers between words and non-words, cf. Appendix \ref{sec:tokenization}) with \texttt{minicons} \cite{misra2022minicons}.

\paragraph{Surprisal} To measure LMs' knowledge of words presented in regular syntactic contexts, we calculate the surprisal of words and non-words $(w,*w)$ as $-log(P(w_i|w_{n<i}))$, i.e. the degree to which the LM is ``surprised'' by the word in the context of plausible preceding tokens, including a BOS token. We create stimuli by sampling sentences that contain our target words from OpenSubtitles \citep{lison2016opensubtitles2016} and substituting them with matching non-words for the false stimuli. If $-log(P(w_i|w_{n<i})) < -log(P(*w_i|w_{n<i}))$, the LM's decision is correct.

\paragraph{Anti-surprisal} Inspired by \citet{shafiabadi2025surprisal}, we include anti-surprisal, a measure of word surprisal on negative instances.
We create negative samples by selecting sentences that our original words do not occur in, and then randomly\footnote{In line with \citet{shafiabadi2025surprisal}, we do not further characterize the resulting violations on a syntactic or semantic level.} placing words/non-words into these sentences at the same index, where $index \geq 3$. By doing so, we compromise between lexical decision and surprisal measurement. There are two reasons to include this measure: i) surprisal in negative samples provides the model with word material as context, but without semantic or syntactic signals that could prime the model towards recognizing it; this allows us to assess whether the mere presence of other words in context makes it easier for the model to distinguish words from non-words, compared to our lexical decision set-up where only whitespace is given. In addition, ii) we want to see if the presence of an ill-fitting context actively deteriorates performance in the sense of a model suddenly preferring non-words over existing words. Again, if $-log(P(w_i|w_{n<i})) < -log(P(*w_i|w_{n<i}))$, the LM's decision is correct.

\paragraph{Learning trajectories}
To assess when word learning happens in relation to syntax learning, we further evaluate intermediate checkpoints of our models on our word learning tests and BLiMP as a syntactic benchmark. In line with previous studies \cite{chang2022word,viering2023shape}, we space our checkpoints logarithmically -- 10 for the first 10\% of training, 9 more for the remaining 90\%. For the Pythia models, we extract similarly spaced checkpoints (the GPT-2 models do not provide checkpoints, so we exclude them).

\section{Results}

\paragraph{Lexical decision} The lexical decision results (Table \ref{tab:results-exp12}) show a strong contrast between character and subword models. Character models achieve near-perfect accuracy (97–99\%) on high-frequency words, regardless of model size. The performance on low-frequency words steadily increases with model size and reaches a near-perfect level for our largest character Llama and the character GPT-2. On the other hand, all BPE models get surprisingly low scores on high-frequency words: The smallest Pythia model discriminates between word and non-words with an accuracy of only 67\%, and the BPE GPT-2 performs below the chance baseline. Even the largest and best BPE model reaches only 87.8\% on high-frequency words -- almost 10\% less than the smallest character model. 

Scaling laws generally hold, with larger models outperforming smaller ones. Interestingly, for the BPE models, there is a consistent gap between high and low-frequency words that cannot be closed by larger models. Smaller character models also show a performance gap between high and low-frequency words, but it narrows considerably with larger models. These results point to substantial differences in how subword and character models learn words. Such a surprising lack of ability in distinguishing words from non-words (without context) is a blatant, hitherto overlooked gap in subword models.

\paragraph{Surprisal and anti-surprisal} Results for the second experiment (Table \ref{tab:results-exp12}) differ from those for lexical decision. In the surprisal setting, the difference between BPE and character models is less pronounced. On high-frequency data, nearly all models (except the smallest BPE Llama) achieve over 90\% accuracy. Still, larger character models yield the best results. In the low-frequency data condition, the pattern is similar, though scores are generally lower. Very large BPE models outperform our Llamas there, but the character GPT-2 remains superior. This may be attributed to the limited lexical exposure of our Llamas, trained on only 10M tokens. In the anti-surprisal setting, character models again drastically outperform BPE models and achieve nearly perfect scores on high-frequency data, while BPE models only reach 70--80\% accuracy. This setting is the only one where Pythia models get better scores on low-frequency data, with a gap of 6--8\%, which increases with model size (not the case for BPE Llamas or character models). This contrast between surprisal and antisuprisal might be an indicator of the entanglement of word learning and syntactic learning in subword models. It is plausible that for high-frequency words, the BPE models have strong expectations about which word should come next in a certain context, and because this expectation is not matched by the real (but ill-fitting) word, a made-up non-word is preferred. We argue that this should still not be the case for an ideal language model -- if a model is indeed well-tuned, it should assign a higher probability to an ill-fitting but existing word which is still in distribution, than to a completely ill-fitting string that is out-of-distribution. In any case, BPE models catch up to character models if (and \textit{only} if) provided with additional syntactic/semantic context information. While random context somewhat aids BPE models, a substantial gap remains between the largest BPE models and the character models, where performance remains excellent, even in the presence of implausible contexts.

\begin{figure}[t!]
\centering
\includegraphics[width=\linewidth]{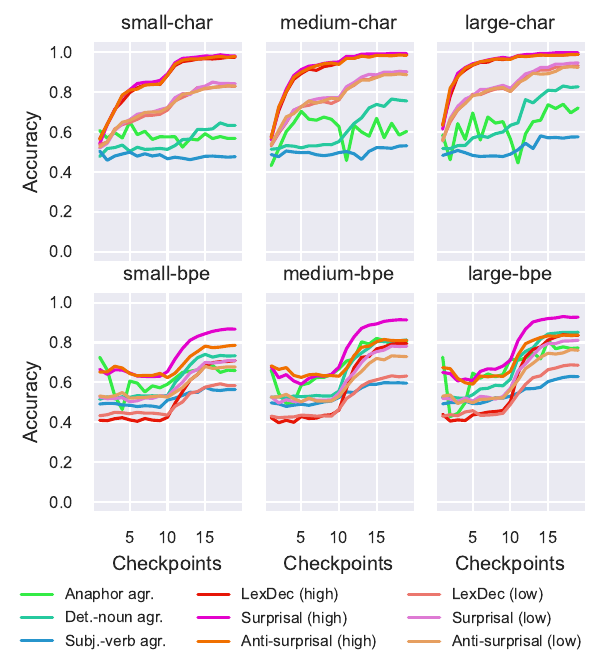}
\vspace{-0.7cm} 
\caption{Selected lexical and syntactic learning curves}
\vspace{-0.5cm} 
\label{fig:linguistic-curves-averaged-n}
\end{figure}

\paragraph{Learning trajectories} Figure \ref{fig:linguistic-curves-averaged-n} displays learning curves for syntactic agreement phenomena, lexical decision, and both surprisal conditions across the 19 saved checkpoints (complete curves reproduced in Appendix \ref{sec:curves}). The first 10 checkpoints correspond to the first 10\% of pretraining, the remaining 9 checkpoints represent 10\% of training steps over the remaining 90\% of pretraining. For character models, the high-frequency, low-frequency, and syntactic curves are clearly separated. On high-frequency data, word learning is rapid and follows power-law curves; the low-frequency scores improve a little later and at a lower rate, but with the same trajectory. Syntactic phenomena improve later, mostly in s-shaped curves (e.g., det.-noun agreement), or are not learned at all in small models (subj.-verb agreement). In contrast, the syntactic and lexical curves for BPE models form sheaves of s-shaped trajectories. There is \textit{no} principal difference in learning dynamics between the syntactic and the word level, improvements\footnote{We calculate Spearman-rank correlation between ordered accuracies for word-level tasks and BLiMP in Appendix \ref{sec:correlations}.} occur simultaneously. This further confirms the results of our previous experiments: in BPE models, word learning is dependent on syntax learning, and words cannot be recognized reliably outside of plausible contexts. Additionally, these different levels of learning cannot be disentangled in BPE models, whereas in character models, syntactic learning \textit{follows} word learning.\footnote {Interestingly, the learning process is not finished when accuracy curves stabilize, cf. Appendix \ref{sec:delta}.}

\section{Discussion}

In contrast to previous work, this study set out to explore when (and if) LMs learn what valid words \textit{are}, and not how LMs learn when words are validly \textit{used}. How should these results now be interpreted in the light of language acquisition and the use of BabyLMs to model corresponding processes? In reality, words are not learned in isolation, but from usage. Yet, words have been widely shown to be represented as solid ``standalone'' units in the mental lexicon, and to be units that human learners acquire early on (cf. \citealp{waxman2009early}, also \citealp{montag2018quantity}). Usage-based approaches have tested aspects of word learning independently from syntax (mostly in object naming tasks, cf. \citealp{tomasello1983joint}) and relate it to pragmatic aspects of communication. For example, children only react to words they know and ignore similar-sounding words. They even struggle with learning words that are phonetically extremely similar (like our stimuli), and only later gain this capacity \cite{stager1997infants}. Similarly, in production, the earliest words come in isolation, slowly emerge into pivot schemas and holophrases \cite{tomasello1992first,tomasello2003constructing}, and only then finally turn into complex sentences. Of course, syntax also aids in discovering  aspects of ``wordiness'', like SV(X)-sentences offering cues for agent-patient relationships. It would be very interesting to further disentangle these levels of word knowledge in a follow-up study, but our current study is not concerned with the ``you shall know a word by the company it keeps''-level of word knowledge, but rather with the ``what do valid words of the language look like''-level, which can be assessed via lexical decision. As such, we believe that the separated learning curves of our character-level models represent more human-like learning than the highly correlated curves found in the subword models, but in reality, an overlap between them is definitely expected.

Reasons for the tremendous performance differences between subword and character models remain open to further inquiry. One plausible explanation is that character models have much more context available to calculate meaningful sequence probabilities, as words are split into many more tokens. While this is true, it is also exactly the point that we are stressing here: it is hard to imagine that arbitrary subword units lead to human-like, plausible word-level representations (like in exemplar models of lexical storage, cf. \citealp{bybee2010language}), whereas character models might offer a more justified level of granularity (and, e.g., better fit reading times, cf. \citealp{oh2021surprisal}). Our findings also align with results on LMs' sensitivity to character-level perturbations \cite{moradi2021evaluating,zhu2024promptbench} and their inability to solve character-level tasks, like counting occurrences of the letter \textit{r} in the word \textit{strawberry} \cite{zhang2024large,shin2024large,cosma2025strawberry}.

\section{Conclusion}

We have shown that the lexical decision approach to the study of word learning in LMs complements surprisal-based approaches and reveals difficulties that surprisal hides: subword LMs struggle with lexical decision, whereas character models master this task with ease. Additionally, in subword LMs, lexical and syntactic learning are inseparable, whereas word learning in character models precedes syntactic learning; the processes are related, yet separable. It is plausible that the \textit{a priori} token splitting in subword models preempts a word discovery process in them, whereas character models first have to pass through this developmental stage, possibly in a somewhat more human-like manner. In any case, as we have shown, decisions about the representational levels of LMs tremendously influence their learning pathways on the different levels of linguistic analysis.

\section*{Limitations}

The generalizability of our findings is constrained by a few factors. The present study has only focused on the English language, but it is plausible that other languages with different writing systems or graphematic and phonotactic rule systems exhibit different patterns under different tokenization schemes. Here, phonetic transcriptions might provide a viable alternative, but real narrow transcriptions that accurately capture the whole breadth of human input are scarce and extremely costly and laborious to manually produce (although novel datasets like \citealp{goriely2025ipachildes} provide an alternative through automatic transcription). Besides, for the character LMs, we focus only on small models, as very large models with such tokenization, especially ones providing intermediate checkpoints, are nonexistent at this moment. It would still be interesting to see how they compare to subword models in a setting where parameter size and training data are greatly increased. 

As already mentioned in Section \ref{sec:relwork}, from a developmental perspective, word learning in humans also includes other processes than statistical pattern recognition from the input: semantic aspects and real-world reference are equally important, as are multimodal input and communicative intent. The ongoing form-vs.-function debate on LMs \cite{mahowald2024dissociating} has begun to consider these aspects, and further studies should aim at incorporating them; for example, the object naming paradigm used in many developmental psychology studies would lend itself naturally to the study of multimodal models.

Finally, we also want to mention that there are attempts to add more linguistic theory to tokenizers. Looking into different tokenizers that, for example, try to be more morphology-aware \cite{hofmann2022embarrassingly,bauwens2024bpeknockout,yehezkel2023incorporating,uzan2024greed} or implement other optimization tricks \cite{schmidt2024tokenization} could yield even more fine-grained points of comparison, but for the present study and its limited scope we focused on the most popular tokenization scheme (BPE) and the most linguistically minimalist alternative (characters).

\section*{Ethical considerations}

Due to the nature of this work, no concrete ethical aspects or repercussions need to be discussed. However, we would like to stress that, of course, BabyLMs not supposed to simulate real babies, but only abstractions of a very specific part of their learning capacity (frequency-driven, domain-general learning mechanisms such as entrenchment or resonance), and therefore all claims about their implications for language development in the real world should be interpreted in this light.

\section*{Acknowledgments}
We would like to thank the three anonymous ARR reviewers for their insightful comments and the highly valuable discussions throughout the rebuttal period.

This research has been funded by the Deutsche Forschungsgemeinschaft (DFG, German Research Foundation) -- CRC-1646, project number 512393437, project A02.

\bibliography{bastian}

\appendix
\newpage
\section{Model hyperparameters and training details}
\label{sec:model-stuff}

\begin{table*}[htbp]
\centering
\footnotesize
\begin{tabular}{@{}l|rrrrrr@{}}
\toprule
 & \texttt{small-char} & \texttt{medium-char} & \texttt{large-char} & \texttt{small-bpe} & \texttt{medium-bpe} & \texttt{large-bpe} \\ \midrule
Embedding size & 128 & 256 & 512 & 128 & 256 & 512 \\
Hidden size & 128 & 256 & 512 & 128 & 256 & 512 \\
Layers & 4 & 8 & 12 & 4 & 8 & 12 \\
Attention heads & 4 & 8 & 12 & 4 & 8 & 12 \\
Context size & 128 & 128 & 128 & 128 & 128 & 128 \\
Vocab. size & 102 & 102 & 102 & 8,002 & 8,002 & 8,002 \\
Parameters & 486,016 & 3,726,592 & 21,940,736 & 2,508,416 & 7,771,392 & 30,030,336 \\ \bottomrule
\end{tabular}
\caption{Model hyperparameters for our self-trained Llama models}
\label{tab:model-params}
\end{table*} 

\begin{figure*}[htb]
\centering
\includegraphics[width=\linewidth]{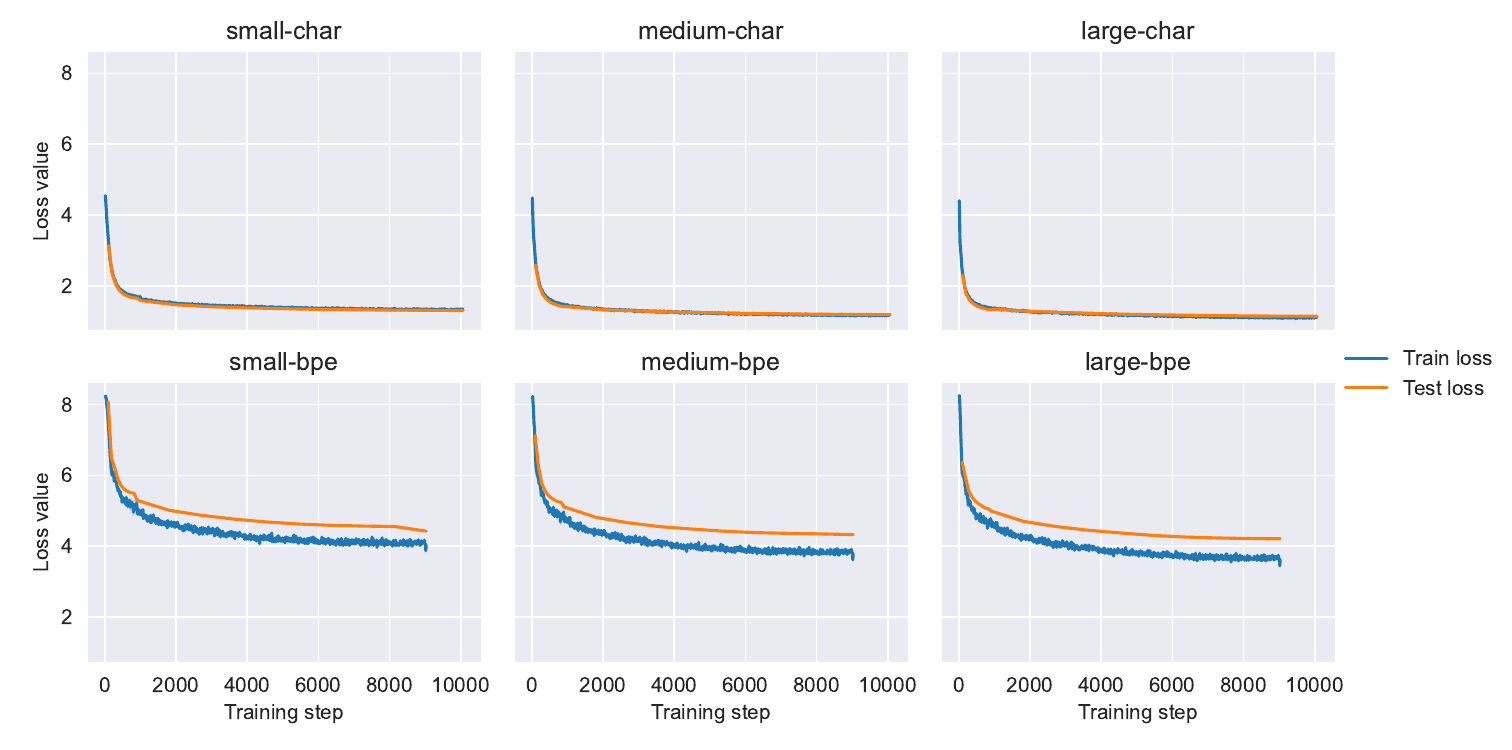}
\caption{Loss curves for our self-trained Llama models}
\label{fig:loss-curves}
\end{figure*}

We use the \texttt{transformers} library \cite{wolf2020transformers} to train our models. The corresponding hyperparameters are listed in Table \ref{tab:model-params}. We opted for very small LMs and little training data because the BabyLM paradigm has consolidated itself as a well-established method of developmentally plausible language modeling \cite{warstadt2023findings,choshen2024call}. Also, we noticed the word learning to occur quite rapidly in our models, so we argue that small LMs offer more fine-grained opportunities for investigating these processes -- in larger LMs, even a singular training step can already influence performance on such a brittle task like lexical decision tremendously.

The subword models feature a considerably higher number of parameters, as the embedding layer of transformer language models accounts for a quite large share of overall model parameters. In light of our vastly different vocabulary sizes (102 for character models, 8,002 for subword models), these differences are not surprising. The subword tokenizer is a regular BPE tokenizer self-trained with the \texttt{tokenizers} library and contains 8,000 subword tokens, a beginning-of-sequence token and a combined end-of-sequence/padding token. The character tokenizer contains these two special tokens and all printable ASCII characters, which are sufficient to represent all graphemes of the English language.

\begin{figure*}[htb]
\centering
\includegraphics[width=0.9\linewidth]{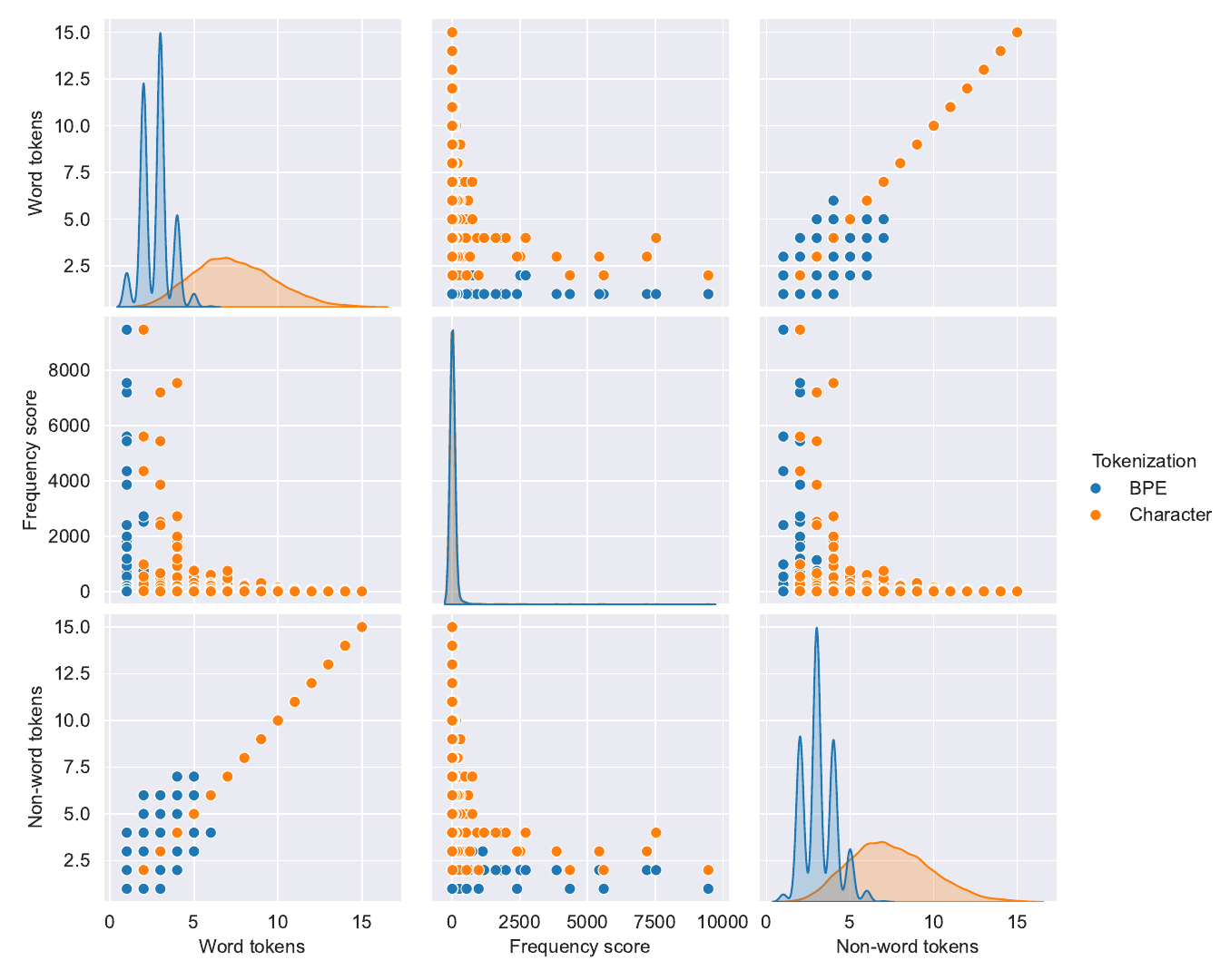}
\caption{Pairplot displaying (i) number of tokens of words, (ii) frequency scores from CELEX \cite{baayen1995celex2} and (iii) number of tokens of non-words (for BPE and character tokenization)}
\label{fig:tokens}
\end{figure*}

Our models were trained on a Apple M2 Pro processor with the MPS backend. For the small models, training lasted approx. 20min, for the medium-sized models it took approx. 1h and for the largest models approx. 10h (no principal differences between character and subword models). We share our models with their checkpoints on the HuggingFace hub.\footnote{\url{https://huggingface.co/collections/bbunzeck/word-learning-in-small-lms-67bdc218688856162b3be08f}}

Loss curves for all models can be found in Figure \ref{fig:loss-curves}. For the test loss, we calculated the perplexity over a held-out portion of our training corpus that is comparable in composition to the training data. We find no principal differences in loss development, although the character models converge faster. Larger models also tend to converge faster and generally reach smaller absolute loss values. As the similar train and test loss curves indicate, all Llama models succeed in optimizing for their next-token prediction objective. However, it remains open to further inquiry how much these scores are constrained by the comparatively small capacity of our BabyLMs, and whether larger models would enhance performance further.

\section{Data creation and tokenization analysis}
\label{sec:tokenization}

\paragraph{Data creation}
Word frequency influences lexical decision performance greatly \cite{mcclelland1981interactive,allen2005evidence}. To incorporate this effect into our study, we create two distinct data sets from words included in \texttt{wuggy}: (i) high-frequency stimuli with a frequency score over 7.0 (occurrences per 1M words in BNC, COCA and other English corpora, as reported in CELEX, \citealp{baayen1995celex2}) and (ii) low-frequency stimuli with a frequency score over 0.0 but below 0.7 (so at least one order of magnitude lower). We opted to rely on the CELEX frequency scores because we compare models trained on different corpora -- our models are trained on the 10M BabyLM corpus, the models by \citet{goriely2024babble} are trained on the 100M BabyLM corpus and the Pythia models are trained on The Pile \cite{gao2020pile}. As such, frequency scores from these corpora would taint analyses of other models and hinder comparability. For the contextualised stimuli, we sample sentences from the OpenSubtitles \cite{lison2016opensubtitles2016} portion of the BabyLM 2024 corpus \cite{choshen2024call}. Both Wuggy and BabyLM data are licensed under the MIT license\footnote{As per license information found at \url{https://github.com/WuggyCode/wuggy} and \url{https://github.com/babylm/evaluation-pipeline-2024}.}, therefore we release our own stimuli artifacts on HuggingFace\footnote{\url{https://huggingface.co/datasets/bbunzeck/lexical-decision}} under the same license. The data contain no information that names or uniquely identifies individual people or offensive content, and are commonly used in computational linguistics.

\begin{figure*}[ht!]
\centering
\includegraphics[width=\linewidth]{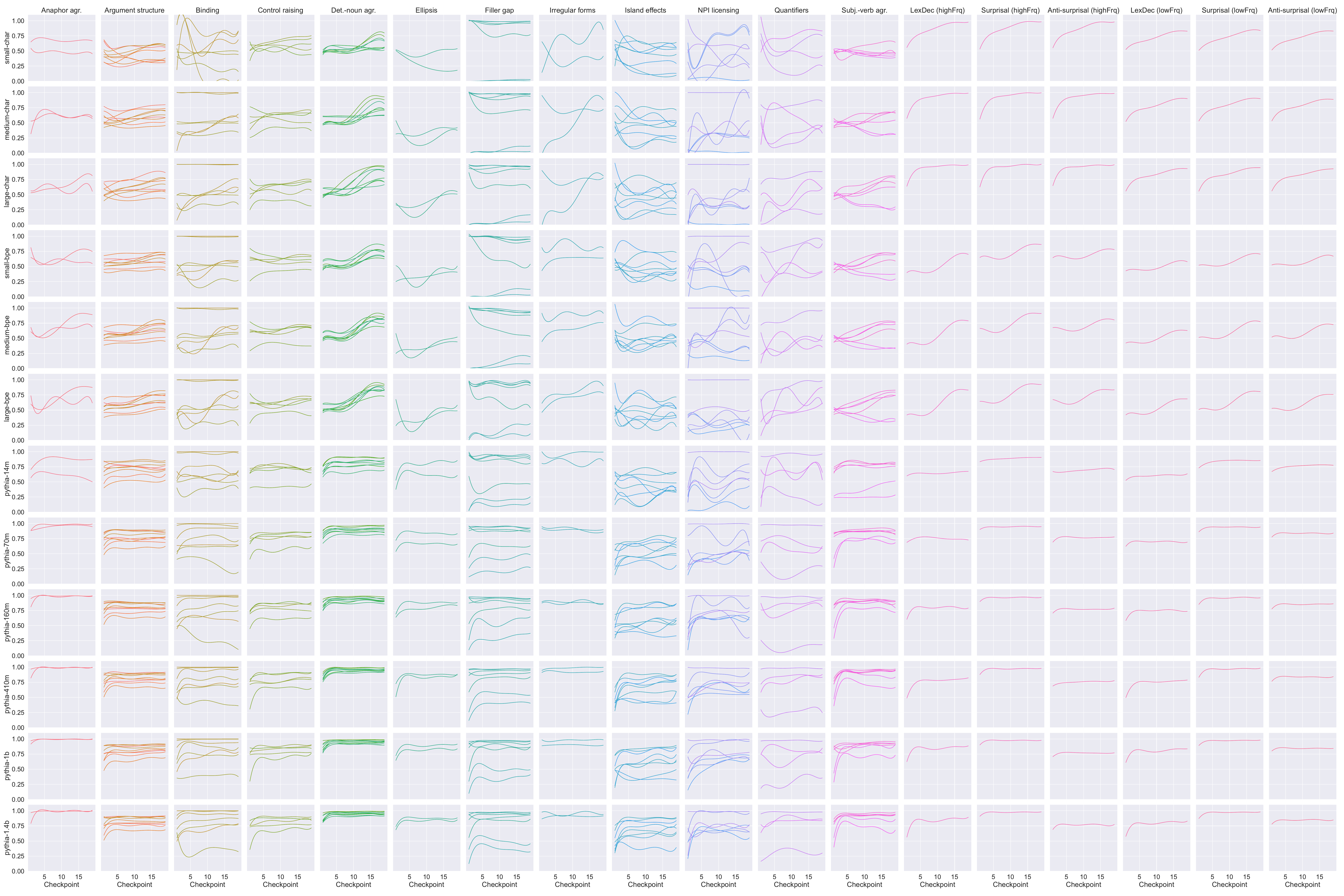}
\caption{Learning curves for all paradigms in BLiMP and high/low frequency lexical decision data, separated for models (rows) and phenomenon sets (columns)}
\label{fig:linguistic-curves}
\end{figure*}

\paragraph{Analysis of tokenization}
To further assess the influence that these frequency scores have on the resulting tokenization for our own models, we offer a brief analysis: Figure \ref{fig:tokens} shows a pairplot between three numerical variables -- (i) the number of tokens that our original words are split into, (ii) the number of tokens that the derived non-word are split into and (iii) the corresponding frequency score from CELEX. While the three plots on the diagonal axis show a layered kernel density estimate (KDE) for each individual variable, the other plots are scatterplots which visualize the relationship between the variables. The data points are colored for their tokenization scheme. 

In the upper left and lower right plot we can see that both real and non-words are split similarly in the two distinct tokenization schemes. Words split by the BPE tokenizer tend to have fewer tokens, mostly between one and six. For the character-based tokenizer, a normal distribution is visible, with its peak at six tokens.

The upper right and lower left scatter plots show the relationship between tokenization for real and non-words. The character-level tokenization exhibits perfect alignment between both kinds of stimuli, they are always split into the exact same number of tokens. Subword tokenization is slightly skewed towards the non-word tokens. This means that non-words are more often split into more tokens than real words, although the reverse case is not completely infrequent.

\section{Full learning curves for BLiMP and word learning}
\label{sec:curves}

Figure \ref{fig:linguistic-curves} shows the full learning curves of all phenomena included in BLiMP (individual syntactic paradigms belonging to one phenomenon are displayed in the same sub-figure) as well as our own lexical benchmarks (all displayed in individual sub-figures), for our six self-trained models and the six Pythia models that we compare them to. We fit a fifth-order polynomial curve to the individual data points and display it on a logarithmic scale. 

It should be noted that we plot the number of the checkpoint on the x-scale. However, the individual amounts of actual textual data seen between these checkpoints differs vastly between our self-trained models (10M lexical tokens) and the Pythia models (825GB of textual data; as the dataset has since been taken down, no lexical token counts are possible anymore).

\section{Correlations between word learning and syntactic learning}
\label{sec:correlations}

\begin{figure*}[ht!]
\centering
\includegraphics[width=0.7\linewidth]{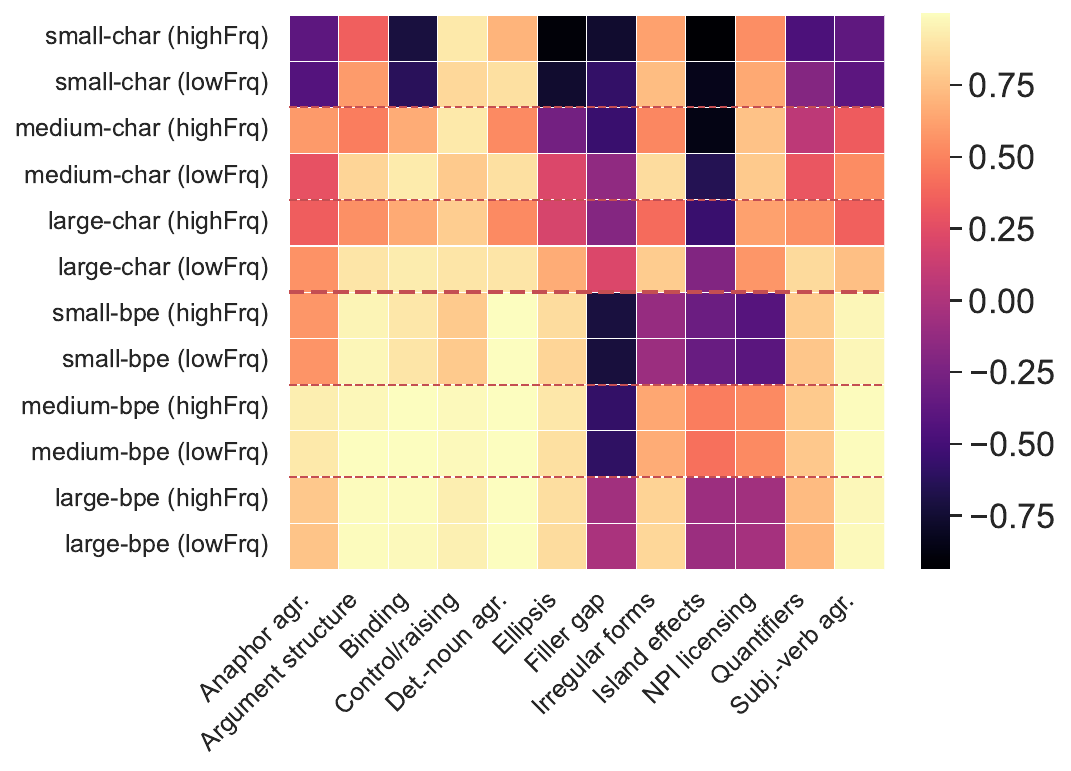}
\caption{Correlation heatmap}
\label{fig:correlation-heat}
\end{figure*}

As an additional measure of commonalities between word learning and syntactic learning, we calculate Spearman-rank correlation scores between ordered accuracy scores for our lexical tasks and BLiMP paradigms. Table \ref{tab:correlations} shows the underlying numerical values for the correlation heatmap provided in Figure \ref{fig:correlation-heat} (please note that the heatmap is rotated in comparison to the table). All scores are statistically significant ($p<0.05$). Due to the similar learning curves found in Figure \ref{fig:linguistic-curves-averaged-n}, we average accuracy scores over all lexical phenomena (lexical decision and both surprisal settings), and then calculate correlations between them (both high and low frequency) and the coarse-grained BLiMP phenomena. For BPE models, lexical performance is highly correlated with more than half of the BLiMP phenomena. The character models show much weaker correlation with syntactic learning. This further confirms our findings about the strong entanglement of lexical and syntactic learning in subword models and their weaker ties in character models.

\begin{table*}[htbp]
\scriptsize
\centering
\begin{tabular}{@{}l|rr|rr|rr|rr|rr|rr@{}}
\toprule
 & \multicolumn{2}{c}{\texttt{small-char}} & \multicolumn{2}{c}{\texttt{medium-char}} & \multicolumn{2}{c}{\texttt{large-char}} & \multicolumn{2}{c}{\texttt{small-bpe}} & \multicolumn{2}{c}{\texttt{medium-bpe}} & \multicolumn{2}{c}{\texttt{large-bpe}} \\ 
BLiMP phenomenon & highFrq & lowFrq & highFrq & lowFrq & highFrq & lowFrq & highFrq & lowFrq & highFrq & lowFrq & highFrq & lowFrq \\ \midrule
Anaphor agr. & -0.393 & -0.435 & 0.580 & 0.277 & 0.332 & 0.555 & 0.569 & 0.559 & 0.930 & 0.910 & 0.772 & 0.759 \\
Argument structure & 0.339 & 0.589 & 0.467 & 0.825 & 0.545 & 0.895 & 0.949 & 0.959 & 0.970 & 0.987 & 0.979 & 0.986 \\
Binding & -0.718 & -0.629 & 0.660 & 0.918 & 0.653 & 0.922 & 0.901 & 0.889 & 0.992 & 0.993 & 0.979 & 0.977 \\
Control raising & 0.904 & 0.837 & 0.909 & 0.776 & 0.791 & 0.892 & 0.777 & 0.780 & 0.974 & 0.974 & 0.930 & 0.936 \\
Det.-noun agr. & 0.686 & 0.870 & 0.524 & 0.869 & 0.521 & 0.890 & 0.989 & 0.989 & 0.990 & 0.993 & 0.994 & 0.989 \\
Ellipsis & -0.912 & -0.766 & -0.285 & 0.209 & 0.180 & 0.662 & 0.857 & 0.822 & 0.897 & 0.868 & 0.865 & 0.856 \\
Filler gap & -0.765 & -0.586 & -0.554 & -0.146 & -0.200 & 0.209 & -0.715 & -0.722 & -0.589 & -0.602 & -0.063 & -0.031 \\
Irregular forms & 0.612 & 0.724 & 0.507 & 0.856 & 0.397 & 0.787 & -0.116 & -0.098 & 0.636 & 0.662 & 0.816 & 0.832 \\
Island effects & -0.937 & -0.840 & -0.862 & -0.657 & -0.556 & -0.214 & -0.321 & -0.334 & 0.473 & 0.418 & -0.088 & -0.094 \\
NPI licensing & 0.535 & 0.641 & 0.748 & 0.782 & 0.616 & 0.568 & -0.425 & -0.407 & 0.520 & 0.526 & -0.069 & -0.049 \\
Quantifiers & -0.476 & -0.202 & 0.059 & 0.299 & 0.547 & 0.848 & 0.789 & 0.767 & 0.779 & 0.770 & 0.716 & 0.695 \\
Subj.-verb agr. & -0.386 & -0.400 & 0.329 & 0.529 & 0.351 & 0.730 & 0.963 & 0.961 & 0.982 & 0.981 & 0.967 & 0.974 \\ \bottomrule
\end{tabular}
\caption{Spearman-rank correlation scores between ordered accuracy scores for our lexical tasks and BLiMP paradigms}
\label{tab:correlations}
\end{table*}

\section{Final BLiMP scores for all models}
\label{sec:full-blimp}

We reproduce the final syntactic evaluation scores for all models that we incorporated in our lexical analyses in Table \ref{tab:results-blimp}. Generally, scores improve with larger models and with more training data. Most strikingly, subword models are consistently superior to comparable character models trained on the same amount of data. These differences, however, are most pronounced for the small models trained on very little data, like our Llama models trained on 10M tokens (7\% for smallest models, 2\% for largest models). For the comparable GPT-2 models trained on 100M tokens, the gap becomes much smaller (0.4\%).

\begin{table}[htb]
\footnotesize
\centering
\begin{tabular}{@{}l|l|r|c@{}}
\toprule
Tok. & Model & Params & BLiMP score \\ \midrule
\multirow{10}{*}{\rotatebox[origin=c]{90}{Subword (BPE)}} & \multirow{6}{*}{Pythia} & 14M & 65.86\%  \\
 &  & 70M & 73.30\% \\
 &  & 160M & 77.50\%  \\
 &  & 410M & 81.63\%  \\
 &  & 1B & 82.21\%  \\
 &  & 1.4B & 81.92\%  \\ \cmidrule(l){2-4}
 & GPT-2 & 85M & 77.80\% \\ \cmidrule(l){2-4}
 & \multirow{3}{*}{Llama} & 2.51M & 59.80\%  \\
 &  & 7.77M & 64.55\%  \\
 &  & 30.03M & 64.56\%  \\ \midrule
\multirow{4}{*}{\rotatebox[origin=c]{90}{Character}} & GPT-2 & 85M & 77.40\% \\ \cmidrule(l){2-4}
 & \multirow{3}{*}{Llama} & 0.49M & 52.69\%  \\
 &  & 3.73M & 51.07\% \\
 &  & 21.94M & 62.14\% \\ \midrule
\end{tabular}
\caption{BLiMP scores for all models}
\vspace{-0.3cm} 
\label{tab:results-blimp}
\end{table}

\section{Development of word/non-word differences}
\label{sec:delta}

\begin{figure*}[ht!]
\centering
\begin{subfigure}{.5\textwidth}
\centering
\includegraphics[width=0.95\linewidth]{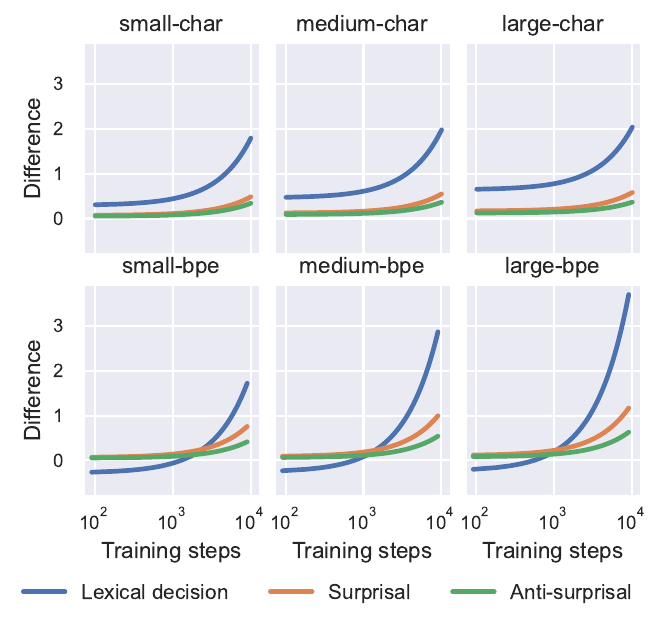}
\caption{High-frequency data}
\label{fig:delta-narrow}
\end{subfigure}%
\begin{subfigure}{.5\textwidth}
\centering
\includegraphics[width=0.95\linewidth]{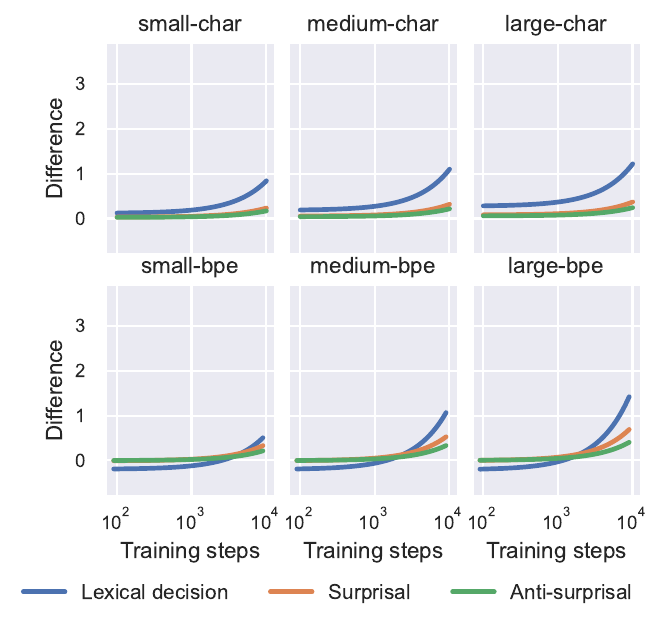}
\caption{Low-frequency data}
\label{fig:delta-curves-low}
\end{subfigure}
\caption{Average differences between surprisal values across pretraining}
\label{fig:delta-curves}
\end{figure*}

In Figure \ref{fig:delta-curves}, we plot the average difference between word and non-word negative log-probability values across training, for both high-frequency and low-frequency data. Positive scores indicate preference for real words. For the character models, the differences are generally less pronounced and get most extreme at the end of pre-training (where accuracy scores do not change anymore), especially for the lexical decision data, which is already consistent at very early training stages. For the BPE models, we see that at the beginning they actually prefer non-words in the lexical decision task. Only after the first 10\% of training they \textit{begin} to discern words and non-words. While overall tendencies remain the same for both frequency conditions, the absolute differences are generally lower and the differences between the curves are less pronounced in the low-frequency setting.

\end{document}